\documentclass[sigconf]{acmart}
\AtBeginDocument{%
  }

\copyrightyear{2026}
\acmYear{2026}
\setcopyright{cc}
\setcctype{by}
\acmConference[MM '26]{Proceedings of the 34th ACM International Conference on Multimedia}{November 10--14, 2026}{Rio de Janeiro, Brazil.}
\acmBooktitle{Proceedings of the 34th ACM International Conference on Multimedia (MM '26), November 10--14, 2026, Rio de Janeiro, Brazil}
\acmDOI{10.1145/3767308.3835042}
\acmISBN{979-8-4007-2213-4/2026/11}

\usepackage{amsmath}
\DeclareMathOperator*{\argmax}{arg\,max}
\DeclareMathOperator*{\argmin}{arg\,min}
\usepackage{colortbl}
\usepackage{booktabs} 
\usepackage{multirow}  
\usepackage{graphicx}

\begin{document}

\title{WIDE: Wildcard Inference with Dynamic Expansion for Cross-Modal Generative Retrieval}

\author{Teng Guo}
\affiliation{%
  \department{College of Computer Science and Technology}
  \institution{Jilin University}
  \city{Changchun}
  \country{China}
}
\email{guoteng24@mails.jlu.edu.cn}

\author{Xin Wang}
\correspondingauthor
\affiliation{%
  \department{College of Computer Science and Technology}
  \institution{Jilin University}
  \city{Changchun}
  \country{China}
}
\additionalaffiliation{%
  \department{Key Laboratory of Symbolic Computation and Knowledge Engineering of Ministry of Education}
  \institution{Jilin University}
  \city{Changchun}
  \country{China}
}
\authornote{The corresponding author.}
\email{w\_x@jlu.edu.cn}

\author{Jiayou Xu}
\affiliation{%
  \department{College of Computer Science and Technology}
  \institution{Jilin University}
  \city{Changchun}
  \country{China}
}
\email{xjy25@mails.jlu.edu.cn}

\author{Keying Zhou}
\affiliation{%
  \department{College of Computer Science and Technology}
  \institution{Jilin University}
  \city{Changchun}
  \country{China}
}
\email{zhouky25@mails.jlu.edu.cn}

\author{Jifeng Shen}
\affiliation{%
  \department{School of Electrical and Information Engineering}
  \institution{Jiangsu University}
  \city{Zhenjiang}
  \country{China}}
\email{shenjifeng@ujs.edu.cn}

\author{Haoxin Ruan}
\affiliation{%
  \department{College of Computer Science and Technology}
  \institution{Jilin University}
  \city{Changchun}
  \country{China}
}
\email{ruanhx24@mails.jlu.edu.cn}
\renewcommand{\shortauthors}{Teng Guo et al.}

\begin{abstract}
Generative retrieval has demonstrated significant success by unifying representation learning and search into a single sequence-to-sequence generation task. However, extending this paradigm to cross-modal retrieval reveals a critical challenge arising from the inherent information asymmetry across different modalities, such as the gap between concise text queries and dense visual candidates. This structural mismatch causes the autoregressive decoder to suffer from forced hallucination when generating identifiers via standard trie-constrained beam search, where the model is severely penalized for failing to guess fine-grained details absent from the query, allowing irrelevant candidates to hijack top rankings. To address this issue, we propose Wildcard Inference with Dynamic Expansion (WIDE). WIDE employs Adaptive Entropy Thresholding (AET) to calibrate layer-specific uncertainty boundaries offline. During the decoding generation phase, Asymmetry-aware Wildcard Decoding (AWD) detects semantic blind spots and emits wildcards instead of forced deterministic identifiers, dynamically expanding the search space without incurring log-probability penalties. Finally, Blind-Spot Re-ranking (BSR) evaluates the expanded candidate pool using a hybrid scoring mechanism that combines discrete generation confidence with continuous semantic similarity. Extensive experiments on the M-BEIR benchmark demonstrate that WIDE outperforms state-of-the-art generative retrieval methods, effectively suppressing forced hallucination while maintaining compact index structures.
\end{abstract}




\maketitle

\section{Introduction}
\label{sec:intro}

\begin{figure}[t!]
    \centering
    \includegraphics[width=0.9\linewidth]{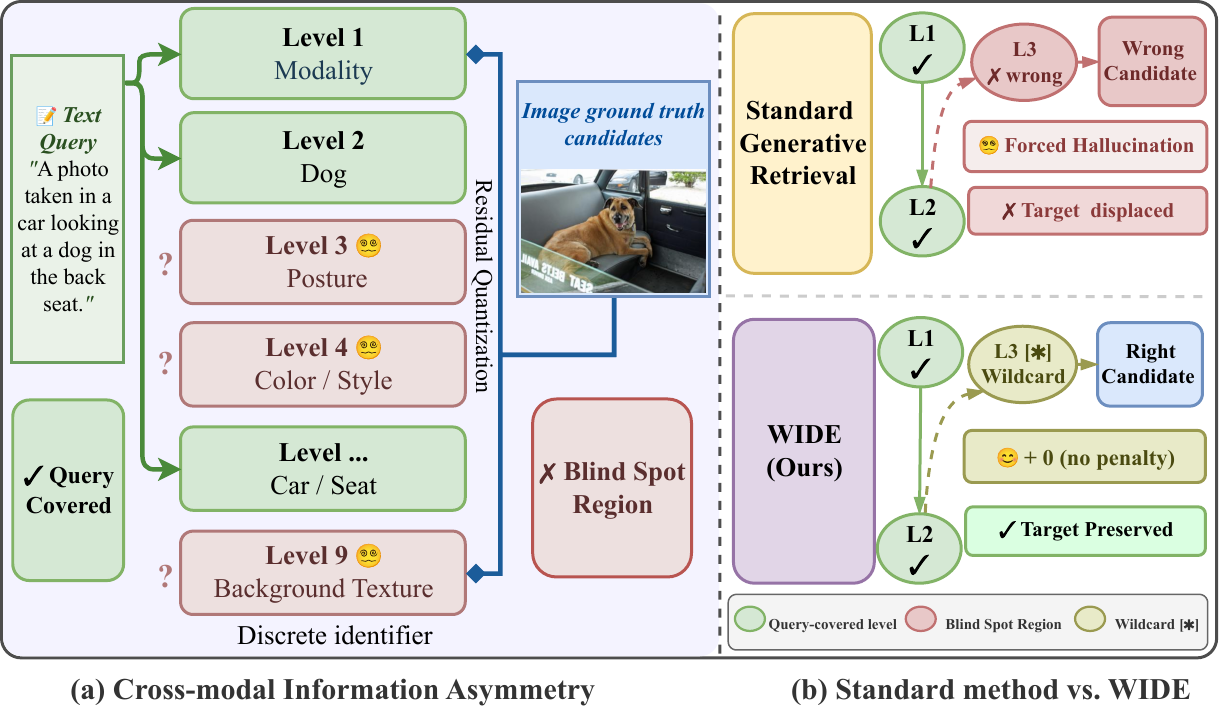}
    \caption{Information asymmetry and methods comparison. Standard beam search forces identifier generation when encountering semantic blind spots. The proposed method detects predictive uncertainty and emits a wildcard identifier.}
    \Description{Panel (a) contrasts the sparse semantic information specified by a text query with the richer information contained in the target image. Panel (b) compares forced identifier generation in standard beam search with WIDE, which emits a wildcard at uncertain identifier levels.}
    \label{fig:information_asymmetry_comparison}
\end{figure}

Generative retrieval introduces a paradigm shift in the architecture of information retrieval systems \cite{tay2022dsi}. Instead of maintaining a fixed dense embedding and performing nearest-neighbor search during inference, this approach formulates retrieval as a sequence-to-sequence generation task \cite{cao2020genre}. By training an autoregressive model to map a query directly to a target identifier, it unifies representation learning and the retrieval process into a single step. This framework has demonstrated significant empirical success in text retrieval settings, where the query and the target share the same modality \cite{tay2022dsi, chen2022corpusbrain}.

However, extending this generative paradigm to cross-modal retrieval reveals a critical limitation rooted in the inherent \emph{information asymmetry} between queries and candidates \cite{liang2022mind}. Current methods process continuous visual features by recursively compressing image embeddings into hierarchical discrete codebook sequences through residual quantization \cite{lee2022autoregressive, zeghidour2021soundstream}. As illustrated in Fig.~\ref{fig:information_asymmetry_comparison}(a), a text query typically outlines only a sparse subset of semantic attributes. In contrast, the candidate image contains an exhaustive record of visual details encoded into these discrete identifiers. This discrepancy reflects the fundamental difference between linguistic abstraction and visual density. Consequently, this structural mismatch disrupts the standard autoregressive decoding process. As illustrated in Fig.~\ref{fig:information_asymmetry_comparison}(b), standard trie-constrained beam search \cite{cao2020genre} forces the decoder to emit a deterministic identifier code at every step. When certain visual details encoded by codebook layers are not specified in the text query, the model encounters semantic blind spots and defaults to statistical co-occurrence patterns learned during training. We define this phenomenon as \emph{forced hallucination}. It constitutes a systematic error driven by cross-modal asymmetry rather than a failure of the underlying representation.

Forced hallucination creates an unfair scoring bias that unnecessarily penalizes correct retrieval paths. When encountering a semantic blind spot, the predictive distribution of the decoder becomes less concentrated, spreading probability mass across multiple identifier candidates~\cite{kuhn2023semantic}. Forcing a deterministic identifier selection under this high uncertainty incurs a severe penalty in the accumulated decoding log-probability. Consequently, a candidate that perfectly matches the query in early layers suffers continuous log-probability drops in uncertain layers simply for failing to guess unmentioned visual details. This flaw allows irrelevant candidates to hijack the top rankings, not through semantic matching, but because their unconstrained details coincidentally align with frequent training priors, while trie constraints only guarantee valid identifier generation rather than relevance.

We argue that this failure mode requires a fundamentally different solution. As illustrated in Fig.~\ref{fig:information_asymmetry_comparison}(b), the objective is not to force the decoder to guess uninformed layers more accurately, as no amount of training can recover information absent from the user query. Instead, the goal is to prevent the decoder from being penalized for inherent information asymmetry. To this end, we propose \textbf{W}ildcard \textbf{I}nference with \textbf{D}ynamic \textbf{E}xpansion for Cross-Modal Generative Retrieval (WIDE). This framework addresses forced hallucination through a cohesive pipeline comprising offline calibration, online intervention, and adaptive candidate evaluation.

The framework operates by first deploying \textbf{A}daptive \textbf{E}ntropy \textbf{T}hresholding (AET) as an offline calibrator to quantify the baseline uncertainty of the model. By profiling the training data, AET establishes layer-specific entropy thresholds that explicitly map the boundaries where semantic blind spots naturally occur. These data-driven boundaries provide a clear reference for the subsequent generation phase. Guided by this reference, \textbf{A}symmetry-aware \textbf{W}ildcard \textbf{D}ecoding (AWD) monitors the real-time autoregressive process to intercept forced hallucinations. Rather than emitting a deterministic token that risks a severe log-probability penalty, AWD generates a wildcard. This specific intervention temporarily suspends the rigid pruning of the prefix trie, dynamically expanding the search space into a semantically coherent candidate set that preserves the correct retrieval path. Finally, to resolve these expanded clusters and prevent irrelevant candidates from hijacking the top results, \textbf{B}lind-\textbf{S}pot \textbf{R}e-ranking (BSR) evaluates the retrieved candidates. BSR ignores unreliable discrete signals from wildcarded positions and combines reliable non-wildcard generation confidence with continuous embedding similarity for final ranking.

Extensive experiments on the M-BEIR benchmark demonstrate that WIDE achieves strong performance compared with existing generative retrieval methods. Further analyses show that wildcard inference with dynamic candidate expansion effectively alleviates forced hallucination while keeping the expanded search space highly constrained.

The main contributions of this paper are summarized as follows:
\begin{itemize}
    \item We identify and formulate \emph{forced hallucination} as a structural failure mode in trie-constrained generative retrieval caused by cross-modal information asymmetry.
    \item We introduce WIDE, a novel framework that addresses this limitation through adaptive uncertainty handling, wildcard-based decoding, and hybrid candidate re-ranking.
    \item We conduct extensive experiments across diverse cross-modal retrieval benchmarks, showing that WIDE achieves state-of-the-art generative retrieval performance and effectively addresses cross-modal information asymmetry.
\end{itemize}

\section{Related Work}
\label{sec:related_work}

\subsection{Generative Retrieval}

Generative Retrieval (GR) \cite{tay2022dsi,wang2022neural} has emerged as an alternative retrieval paradigm that replaces the conventional match-and-rank pipeline with direct identifier generation. 

Building upon success in text retrieval, GR has been extended to cross-modal scenarios \cite{li2024generative,qu2024tiger,li2025revolutionizing,li2025semcore,fang2025cart}. Unlike single-modal retrieval~\cite{li2025matching}, cross-modal retrieval must address the representational gap between heterogeneous modalities, which complicates the design of identifiers. GENIUS \cite{kim2025genius} proposes a generalized multimodal generative retrieval framework. The Modality-Decoupled Semantic Quantization of this framework learns modality-invariant semantic anchors to improve identifier generation across different modalities. While this line of work emphasizes the semantic quality and modality alignment of identifiers, existing methods largely overlook the inherent information asymmetry between queries and visual targets during the decoding process.

Furthermore, hybrid designs incorporating late fusion \cite{song2024re3val} have recently been introduced to generative retrieval. Representative approaches ~\cite{yang2024unifying,yang2025sparse} generate initial candidate identifiers and subsequently refine them using additional similarity signals. These designs are highly practical, as they preserve the efficiency of discrete generative retrieval while leveraging complementary continuous signals to correct the outputs of discrete generation. However, current reranking strategies generally apply flat embedding similarity metrics, failing to account for the specific semantic blind spots caused by decoding uncertainty.

\subsection{Constrained Decoding and Uncertainty Estimation}

Trie-constrained beam search \cite{hokamp2017lexically,anderson2017guided,cao2020genre} serves as the standard decoding strategy in GR. 
This mechanism organizes all valid target sequences into a prefix tree, where each path from the root to a leaf corresponds to a registered identifier. 
While this hard constraint prevents the generation of out-of-vocabulary identifiers, it forces the decoder to make a specific identifier decision even when the query provides insufficient evidence.

Entropy-based uncertainty estimation has been extensively explored in generative modeling \cite{xia2025survey}. Prior research \cite{malinin2020uncertainty} demonstrates that predictive entropy serves as an effective signal for reliability estimation at both the token and sequence levels. Recent studies \cite{kuhn2023semantic,zhang2025tokur} propose semantic entropy to capture uncertainty beyond variations in surface forms, while token-level uncertainty enables fine-grained confidence assessment and factuality evaluation. 

These findings indicate that uncertainty signals derived from the output distribution of the model provide critical guidance for identifying unreliable generation. 
Our work leverages this metric to dynamically calibrate decoding interventions, transforming unguided forced guesses into structured candidate expansion.

\begin{figure*}[t]
    \centering
    \includegraphics[width=1.0\linewidth]{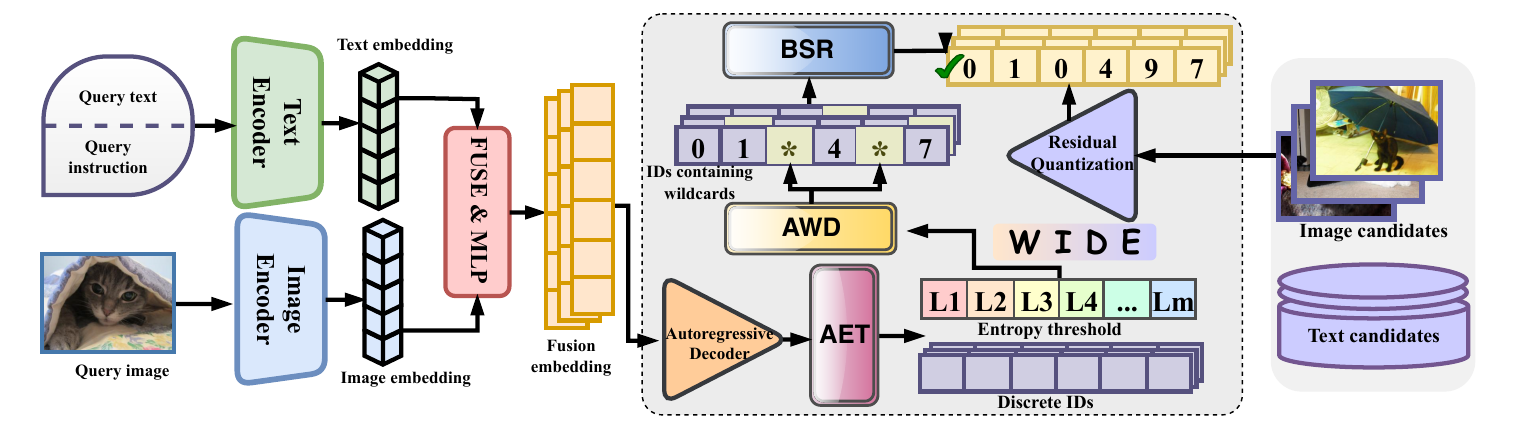}
    \caption{Overview of the proposed WIDE framework. Within the generative inference pipeline (dashed box), AET establishes layer-wise entropy thresholds to guide AWD in dynamically emitting wildcards at semantic blind spots. BSR subsequently evaluates the wildcard-expanded candidate sets to yield the retrieval results.}
    \Description{A pipeline diagram of WIDE. Query and candidate encoders produce representations used for residual quantization and identifier generation. AET supplies adaptive layer-specific entropy thresholds, AWD emits wildcards at uncertain levels, and BSR ranks the expanded candidates.}
    \label{fig:wide_framework_overview}
\end{figure*}

\section{Preliminary}
\label{sec:preliminary}

\subsection{Problem Formulation}

We consider cross-modal generative retrieval over a heterogeneous candidate database $\mathcal{C} = \{c^{(1)}, \dots, c^{(N)}\}$, where each candidate $c$ represents a multimodal entity. A composite query $\mathbf{q}$ comprises content signals, including text, images, and interleaved modalities, alongside task-specific instructions that guide the desired target retrieval \cite{wei2024uniir}. 

Generative retrieval formulates this task as a constrained sequence generation process \cite{tay2022dsi}. Each candidate $c \in \mathcal{C}$ is uniquely mapped to a discrete identifier sequence $T_c = (c_1, \dots, c_M)$. The retrieval objective is to maximize the conditional generation probability of the identifier sequence given the query context according to the following formulation.

\begin{equation}
    \hat{c} = \argmax_{c \in \mathcal{C}} \prod_{k=1}^{M} p_\theta\!\left(c_k \mid \mathbf{q},\, c_{<k}\right),
    \label{eq:gen_ret}
\end{equation}

\noindent where $c_{<k}$ denotes the generated sequence prefix and $\theta$ represents the model parameters. To ensure validity, the generated sequence must traverse a pre-constructed prefix trie containing all registered database identifiers \cite{cao2020genre}. This structural constraint enables the system to bypass exhaustive similarity scoring across the entire database.

The effectiveness of this paradigm relies on two interconnected stages. It first requires learning compact and discriminative identifiers during quantization and then accurately generating these identifiers during decoding, especially when queries provide incomplete semantic coverage. Our work focuses on addressing the decoding-stage challenge.

\subsection{Residual Quantization}
\label{sec:prelim_rq}

To map multimodal representations into the discrete space, Residual Quantization (RQ) compresses a dense candidate embedding $\mathbf{v}_c \in \mathbb{R}^D$ through an iterative residual approximation~\cite{lee2022autoregressive, zeghidour2021soundstream}. Given a sequence of $M$ codebooks $\{\mathcal{B}_1, \dots, \mathcal{B}_M\}$, where each $\mathcal{B}_k = \{\mathbf{b}_{k,v} \in \mathbb{R}^D\}_{v=1}^{V_k}$ contains $V_k$ code vectors, the quantization process initializes the residual vector as $\mathbf{\Delta}_1 = \mathbf{v}_c$. In our implementation, we use $M=9$ quantization levels, where the first level encodes modality information. At each quantization level $k \in \{1,\dots,M\}$, RQ identifies the nearest code vector to the current residual state:
\begin{equation}
    c_k = \argmin_{v \in [V_k]} \left\|\mathbf{\Delta}_k - \mathbf{b}_{k,v}\right\|_2^2, \qquad 
    \mathbf{\Delta}_{k+1} = \mathbf{\Delta}_k - \operatorname{sg}\!\left(\mathbf{b}_{k,c_k}\right),
    \label{eq:rq}
\end{equation}
\noindent where $[V_k]=\{1,\dots,V_k\}$ and $\operatorname{sg}(\cdot)$ denotes the stop-gradient operator. The resulting quantized approximation is $\hat{\mathbf{v}}_c=\sum_{k=1}^{M}\mathbf{b}_{k,c_k}$.

The codebooks are initialized using $k$-means and updated through exponential moving averages. The RQ objective encourages the residual representation at each level to remain close to its selected code vector:
\begin{equation}
    \mathcal{L}_{\mathrm{RQ}}=
    \frac{1}{MD}\sum_{k=1}^{M}
    \left\|
    \mathbf{\Delta}_k-\operatorname{sg}\!\left(\mathbf{b}_{k,c_k}\right)
    \right\|_2^2.
    \label{eq:rq_loss}
\end{equation}

By sequentially minimizing the residual error, Eq.~\eqref{eq:rq} provides an ordered refinement of the quantized representation. The shallow levels capture dominant components of the candidate representation, while deeper levels encode the remaining residual information, which often corresponds to finer-grained visual details.

This hierarchical residual structure introduces a challenge during cross-modal retrieval. Textual queries are inherently selective and sparse. They usually specify particular objects or attributes while omitting visual details that are not explicitly mentioned. Consequently, a query may provide strong semantic constraints for some identifier levels while leaving other levels insufficiently grounded. This varying degree of query coverage across quantization levels can disrupt the subsequent identifier generation process.

\subsection{Trie-Constrained Decoding and Forced Hallucination}
\label{sec:prelim_decoder}

To perform retrieval, an autoregressive decoder generates the target sequence $T_c$ conditioned on the query. The model is optimized via token-level cross-entropy using the ground-truth sequence $\mathbf{c}^*$:
\begin{equation}
    \mathcal{L}_\text{GR} = -\sum_{k=1}^{M} \log p_\theta\!\left(c^*_k \mid \mathbf{q},\, c^*_{<k}\right).
    \label{eq:decoder_loss}
\end{equation}

During inference, beam search is employed to explore the prefix trie. To guarantee that the output corresponds to an existing candidate, the token space at step $k$ is restricted to $\mathcal{T}_k$, which represents the admissible set of valid child nodes extending from the current prefix $c_{<k}$~\cite{cao2020genre}. The beam search algorithm evaluates hypotheses based on their cumulative log-probability:
\begin{equation}
    \text{Score}(T_c) = \sum_{k=1}^{M} \log p_\theta\!\left(c_k \mid \mathbf{q},\, c_{<k}\right).
    \label{eq:beam_score}
\end{equation}

While the constraint space $\mathcal{T}_k$ prevents the generation of out-of-vocabulary identifiers, it requires the decoder to commit to a specific identifier choice at every step. Under cross-modal information asymmetry, when the decoder reaches a quantization level where the query provides insufficient semantic evidence, the predictive distribution $p_\theta(t \mid \mathbf{q}, c_{<k})$ may become less concentrated and dominated by prefix-dependent statistical priors rather than query-specific evidence~\cite{kuhn2023semantic}.

Despite this uncertainty, standard trie-constrained decoding forces the model to select a specific token. This forced selection can incur a log-probability penalty~\cite{stahlberg2019nmt}, approaching $-\log|\mathcal{T}_k|$ when the distribution becomes nearly uniform. We define this behavior as \emph{forced hallucination}. A hypothesis beam that correctly captures the query-informed attributes in early layers will accumulate massive numerical penalties in the unconstrained layers. Consequently, the ground-truth candidate is frequently displaced by irrelevant distractors whose deeper identifiers happen to statistically align with the training priors of the decoder~\cite{holtzman2019curious}. Addressing this structural failure necessitates a dynamic intervention mechanism that quantifies query coverage boundaries, permits explicit expression of uncertainty, and adaptively evaluates the resulting candidate sets. We detail our proposed framework to achieve these objectives in Section~\ref{sec:method}.

\section{Methodology}
\label{sec:method}

\subsection{Overview}
\label{sec:method_overview}

Generative retrieval recasts the traditional dense matching paradigm into an autoregressive sequence generation task. As illustrated in Fig.~\ref{fig:wide_framework_overview}, our pipeline processes multimodal inputs through specific encoders. On the query side, the query text, task instructions and the query image are encoded and subsequently integrated via a fusion module (FUSE \& MLP) to form a unified fusion embedding. On the candidate side, database images and texts are mapped into a shared semantic space, where RQ compresses them into hierarchical discrete identifiers. 
During training, an autoregressive decoder learns to generate the target discrete identifier conditioned on the unified fusion embeddings.

To resolve forced hallucination highlighted above, we propose the WIDE framework. As depicted in the dashed box of Fig.~\ref{fig:wide_framework_overview}, WIDE integrates into the generative retrieval inference pipeline through three sequential modules. First, AET (Sec.~\ref{sec:method_aet}) computes layer-wise entropy thresholds ($L_1 \dots L_m$) from the training data to serve as an offline reference for predictive uncertainty. Next, AWD (Sec.~\ref{sec:method_awd}) utilizes these entropy thresholds to monitor the autoregressive decoding process. When predictive uncertainty surpasses the entropy threshold, it bypasses forced generation by inserting a wildcard into the search path, seamlessly producing discrete identifier sequences without terminating the generation. Finally, BSR (Sec.~\ref{sec:method_bsr}) evaluates the candidate pool expanded by AWD using a hybrid scoring mechanism to yield the final retrieval results.

\subsection{Adaptive Entropy Thresholding}
\label{sec:method_aet}

The central premise of AWD is that predictive uncertainty of the output distribution at level $k$ provides a useful signal for identifying potential query semantic blind spots caused by cross-modal information asymmetry. However, raw entropy values are not directly comparable across levels because the trie-constrained admissible set size $|\mathcal{T}_k|$ varies across RQ levels and prefixes, which affects the entropy scale. Therefore, evaluating uncertainty requires a level-specific, data-driven reference. AET establishes this reference offline to calibrate the online decoding process.

\begin{figure}[t]
    \centering
    \includegraphics[width=1\linewidth]{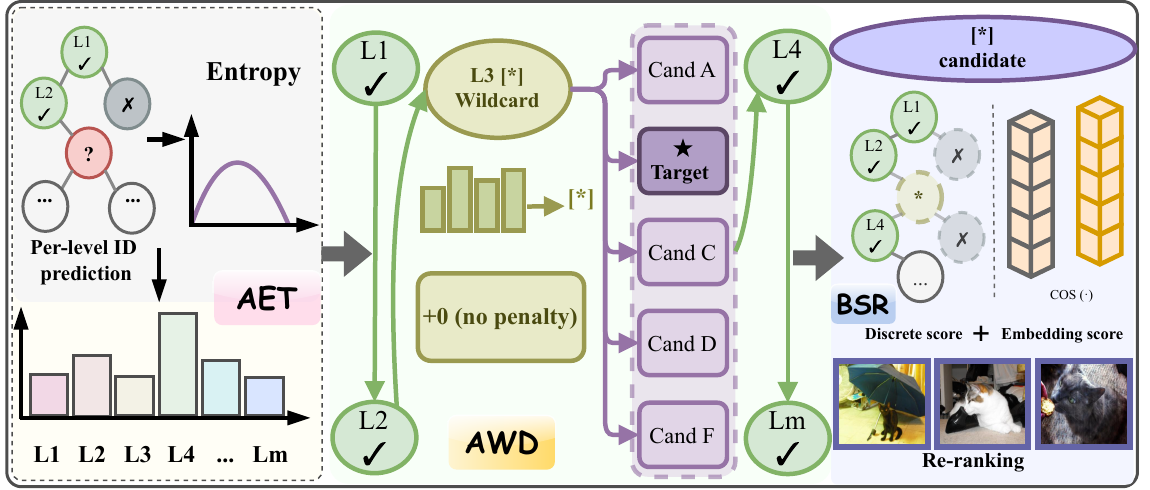}
    \caption{The architecture of AET, AWD and BSR. (1) AET quantifies predictive uncertainty by calculating the entropy of per-level identifier generation to establish thresholds. (2) AWD circumvents forced generation by inserting a wildcard ($\ast$) with no penalty, maintaining the continuous expansion of the sequence. (3) BSR re-ranks the wildcard-inclusive candidate sequences by fusing discrete structural scores with embedding similarities.}
    \Description{Architecture of AET, AWD, and BSR in WIDE. AET estimates level-specific entropy thresholds, AWD replaces unreliable identifier generations with wildcards, and BSR combines reliable generation confidence with embedding similarity for candidate re-ranking.}
    \label{fig:aet_awd_bsr_architecture}
\end{figure}

Given a generative retrieval model \(p_\theta\) and the training data \(\{(\mathbf q,\mathbf c^*)\}\), as shown in the left panel of Fig.~\ref{fig:aet_awd_bsr_architecture}, we define the dynamic entropy threshold \(\tau_k\) as the expected predictive entropy at each level \(k\) under teacher forcing. By conditioning the decoder on the true prefix \(c^*_{<k}\), we estimate the predictive uncertainty at each level under a consistent generation context:
\begin{equation}
    \tau_k = \bar{H}_k = \mathbb{E}_{(\mathbf{q},\,\mathbf{c}^*)}
    \left[ 
        -\sum_{t \in \mathcal{T}_k} 
        p_\theta\!\left(t \mid \mathbf{q},\, c^*_{<k}\right) 
        \log p_\theta\!\left(t \mid \mathbf{q},\, c^*_{<k}\right)
    \right].
    \label{eq:threshold_entropy}
\end{equation}
This expected entropy \(\bar H_k\) captures the typical uncertainty associated with each RQ level. Shallow levels often exhibit lower uncertainty because they encode dominant residual components, whereas deeper levels may exhibit higher uncertainty when the remaining visual details are insufficiently specified by the query.

Using offline expected entropy \(\bar{H}_k\) as a level-specific reference for real-time entropy \(H_k\) allows AWD to identify unusually uncertain decoding states~\cite{hendrycks2017baseline}. During inference, if a novel query yields a real-time entropy \(H_k > \tau_k\), it indicates that the decoder is more uncertain than the historical baseline for that specific semantic level. This parameter-free comparison enables adaptive wildcard triggering while accounting for the intrinsic uncertainty differences across RQ levels.

\subsection{Asymmetry-aware Wildcard Decoding}
\label{sec:method_awd}

Standard trie-constrained beam search maintains \(B\) identifier prefixes and selects the top-\(B\) valid extensions at each level according to cumulative log-probability. When the decoder encounters a semantic blind spot, it is still forced to select a specific identifier code from the admissible set \(\mathcal{T}_k\). Under high uncertainty, this forced decision introduces a large negative log-probability contribution that can suppress the entire retrieval path. AWD avoids this unnecessary penalty by replacing deterministic identifier generation at uncertain levels with wildcards.

At each decoding level $k$, AWD computes the entropy of the decoder output distribution over $\mathcal{T}_k$.
\begin{equation}
    H_k = -\sum_{t \in \mathcal{T}_k}
    p_\theta\!\left(t \mid \mathbf{q},\, c_{<k}\right)
    \log p_\theta\!\left(t \mid \mathbf{q},\, c_{<k}\right),
    \label{eq:entropy}
\end{equation}
As shown in the middle panel of Fig.~\ref{fig:aet_awd_bsr_architecture}, AWD applies a binary decision for each active beam $b$. When $H_k \leq \tau_k$, the query provides sufficient constraint, and the log-probability of the selected token is added to the beam score. When $H_k > \tau_k$, AWD registers level $k$ as a blind spot, records it in the specific blind-spot set $\mathcal{A}_b$ of the beam, and emits a wildcard $[\ast]$. Crucially, no log-probability penalty is added to the beam score for this level.
\begin{equation}
    \log P^{(AWD)}(T_b) =
    \sum_{k \notin \mathcal{A}_b}
    \log p_\theta\!\left(c_k \mid \mathbf{q},\, c_{<k}\right),
    \label{eq:awd_score}
\end{equation}
This scoring rule ensures that each beam is evaluated only on the levels where the identifier generation is supported by the query.

Instead of forcing an unreliable identifier code generation, the prefix trie temporarily activates all valid child branches when a wildcard is triggered. However, AWD does not terminate the autoregressive decoding at this first semantic blind spot. If the query provides explicit semantic constraints at a subsequent deeper level ($k \notin \mathcal{A}_b$), the decoder seamlessly resumes standard identifier code generation to prune the expanded branches. Let $\hat{c}_k$ denote the confident identifier code generated at non-blind-spot levels. The final candidate set retrieved by beam $b$ is defined by dynamic set containment.
\begin{equation}
    \mathcal{S}_b = \left\{
        \mathbf{c} \in \mathcal{C}
        \;\middle|\;
        c_k = \hat{c}_k \;\; \forall\, k \notin \mathcal{A}_b
    \right\},
    \label{eq:subtree}
\end{equation}
The set $\mathcal{S}_b$ contains every database candidate that perfectly matches the confident discrete generations of the beam, while leaving the wildcard levels unconstrained. Thus, intermediate blind spots are explicitly bypassed rather than forcibly generated, allowing reliable identifier constraints from different levels to be jointly preserved without prematurely abandoning the search.

The dynamic expansion mechanism may introduce additional candidates. Under a uniform identifier distribution assumption, activating a wildcard after fixing the first $m-1$ identifier levels reduces the expected candidate space by a factor of $V^{m-1}$, yielding an expected candidate size of $|\mathcal{S}_b| \approx |\mathcal{C}|/V^{m-1}$. This suggests that wildcard expansion remains limited when reliable identifier prefixes are preserved. Since real identifier distributions may be non-uniform, we further analyze the actual expansion behavior empirically.

At the end of the AWD process, all candidates satisfying the wildcard constraints of the retained beams form a global candidate pool \(\mathcal{S}_{global}\), which is then forwarded to BSR.

\subsection{Blind-Spot Re-ranking}
\label{sec:method_bsr}

Following the AWD process, the top-$B$ beam paths yield a global candidate pool $\mathcal{S}_{global}$. Candidates associated with the same beam share the same reliable identifier constraints at non-wildcard levels, while candidates from different beams correspond to different decoding paths. Relying solely on continuous similarity to rank $\mathcal{S}_{global}$ is suboptimal because it ignores the discrete identifier constraints preserved during generation and treats wildcard-induced variations equally. To address this issue, BSR introduces a hybrid scoring function that combines reliable discrete generation confidence from constrained levels with continuous embedding similarity.

As shown in the right panel of Fig.~\ref{fig:aet_awd_bsr_architecture}, for each candidate $c^{(i)} \in \mathcal{S}_{global}$ originating from a beam with a blind-spot set $\mathcal{A}_i$, the final retrieval score $s(i)$ is defined as a convex combination of the discrete and continuous scores.
\begin{equation}
    s(i) =
    (1 - \alpha_i) \cdot \frac{1}{M - |\mathcal{A}_i|} \sum_{k \notin \mathcal{A}_i} p_\theta\!\left(c_k^{(i)} \mid \mathbf{q},\, c_{<k}^{(i)}\right)
    +\alpha_i \cdot \frac{1 + \cos\!\left(\mathbf{f}_q,\, \mathbf{f}_i\right)}{2},
    \label{eq:bsr}
\end{equation}
where $\mathbf{f}_q$ and $\mathbf{f}_i$ denote the query and candidate embeddings used to compute continuous semantic similarity.

The first term computes the average autoregressive probability over all levels that are not identified as blind spots ($k \notin \mathcal{A}_i$). Because AWD skips unreliable levels while continuing generation at the remaining levels, confident identifier generations are distributed across the sequence. Therefore, averaging over the valid sequence length ($M-|\mathcal{A}_i|$) normalizes the discrete score to the $[0,1]$ range. This normalization neutralizes the length bias caused by varying blind-spot proportions across different beams. This mechanism explicitly utilizes the identifier generation confidence of the model to rank candidates based on query. The second term computes the cosine similarity between the global embeddings, shifted to the $[0, 1]$ range to align with the scale of the discrete probabilities.

To regulate the contribution of each term without relying on empirical hyperparameters, the interpolation weight $\alpha_i$ is determined by the proportion of blind spots in the source beam of the candidate.
\begin{equation}
    \alpha_i = \frac{|\mathcal{A}_i|}{M},
    \label{eq:alpha}
\end{equation}
where $M$ is the maximum sequence length. When a query provides detailed constraints ($|\mathcal{A}_i|$ is small), $\alpha_i$ approaches 0, and the ranking primarily depends on the discrete generative confidence. When the query lacks specific details and triggers multiple wildcards ($|\mathcal{A}_i|$ is large), $\alpha_i$ increases. This shift assigns greater weight to the continuous semantic similarity, utilizing the global cross-modal embeddings to resolve visual ambiguities within the expanded clusters. Finally, BSR sorts $\mathcal{S}_{global}$ by $s(i)$ to output the top-$K$ retrieval results.

\section{Experiments}

To validate the effectiveness of the proposed WIDE framework, we conduct extensive experiments focusing on its ability to resolve semantic blind spots under cross-modal asymmetry. We benchmark our method against state-of-the-art embedding-based and generative retrieval paradigms.

\subsection{Experimental Setup}
\label{sec:exp_setup}

\textbf{Datasets and Evaluation Metrics.}
We evaluate our model on the comprehensive M-BEIR benchmark~\cite{wei2024uniir}, which aggregates 10 datasets and a pool of 5.6 million candidate items across eight distinct multimodal retrieval tasks. Specifically, the benchmark encompasses widely adopted datasets including MS-COCO \cite{lin2014microsoft}, VisualNews \cite{liu2021visualnews}, Fashion200K \cite{han2017automatic}, NIGHTS \cite{fu2023dreamsim}, FashionIQ \cite{wu2021fashion}, CIRR \cite{liu2021cirr}, WebQA \cite{chang2022webqa}, OVEN \cite{hu2023oven}, InfoSeek \cite{chen2023can}, and EDIS \cite{liu2023edis}. Rather than a monolithic collection, M-BEIR encompasses various complex scenarios: standard cross-modal retrieval, fine-grained attribute manipulation, image-to-image similarity reasoning. Adhering strictly to the established M-BEIR evaluation protocol, we report Recall@5 (R@5) as the primary evaluation metric across most tasks, and adopt Recall@10 (R@10) specifically for Fashion200K and FashionIQ to ensure fair comparisons.

\textbf{Baseline.}
We acknowledge the foundational contributions of the GENIUS framework~\cite{kim2025genius}, as its multimodal general generative retrieval architecture serves as the direct starting point for our approach. Consequently, we establish a primary baseline that adopts a highly analogous two-stage architecture relying exclusively on residual quantization and autoregressive decoding. Our baseline utilizes the fine-tuned variants of CLIP with score-level fusion (SF)~\cite{radford2021learning}, as the foundational dual-encoder to extract visual and textual representations. For the generative component, we instantiate the autoregressive decoder using a randomly initialized T5-small model~\cite{raffel2020exploring}. Due to computational limitations, we train this baseline model using a constrained hardware configuration consisting of two NVIDIA L20 GPUs.

\textbf{Training Strategy.}
The network is optimized using the AdamW optimizer~\cite{loshchilov2017decoupled} with a batch size of 256. We apply a peak learning rate of $1 \times 10^{-4}$ coupled with a cosine decay schedule. The RQ module is trained for 20 epochs, and the autoregressive decoder is subsequently trained for 30 epochs.

\subsection{Overall Retrieval Performance}
\label{sec:main_results}

\begin{table*}[t]
    \centering
    \caption{Comprehensive Performance on the M-BEIR Benchmark. Unlike previous works, we group the diverse task formulations by their target modality to demonstrate the robust generalization of our method. Metrics include Recall@5 and Recall@10.}
    \label{tab:overall_performance}
    \resizebox{\textwidth}{!}{
    \begin{tabular}{l ccc c cc c c ccc cc c cc cc}
        \toprule
        \multirow{3}{*}{\textbf{Method}} & 
        \multicolumn{6}{c}{\textbf{Target: Image Retrieval ($c_i$)}} & &
        \multicolumn{6}{c}{\textbf{Target: Text Retrieval ($c_t$)}} & &
        \multicolumn{4}{c}{\textbf{Target: Multimodal ($c_i, c_t$)}} \\
        
        \cmidrule{2-7} \cmidrule{9-14} \cmidrule{16-19}
        
        & \multicolumn{3}{c}{Query: $q_t$} & \multicolumn{1}{c}{$q_i$} & \multicolumn{2}{c}{$(q_i, q_t)$} & &
        \multicolumn{1}{c}{Query: $q_t$} & \multicolumn{3}{c}{$q_i$} & \multicolumn{2}{c}{$(q_i, q_t)$} & &
        \multicolumn{2}{c}{Query: $q_t$} & \multicolumn{2}{c}{$(q_i, q_t)$} \\
        
        \cmidrule(lr){2-4} \cmidrule(lr){5-5} \cmidrule(lr){6-7} 
        \cmidrule(lr){9-9} \cmidrule(lr){10-12} \cmidrule(lr){13-14} 
        \cmidrule(lr){16-17} \cmidrule(lr){18-19}
        
        & COCO & VN & F200K & NIGHTS & FIQ & CIRR & & WebQA & COCO & VN & F200K & OVEN & InfoS & & EDIS & WebQA & OVEN & InfoS \\
        & {\scriptsize R@5} & {\scriptsize R@5} & {\scriptsize R@10} & {\scriptsize R@5} & {\scriptsize R@10} & {\scriptsize R@5} & & {\scriptsize R@5} & {\scriptsize R@5} & {\scriptsize R@5} & {\scriptsize R@10} & {\scriptsize R@5} & {\scriptsize R@5} & & {\scriptsize R@5} & {\scriptsize R@5} & {\scriptsize R@5} & {\scriptsize R@5} \\
        \midrule
        
        \multicolumn{19}{c}{\textit{Embedding-based Retrieval}} \\
        \midrule
        CLIP-SF & 81.1 & 42.6 & 18.0 & 32.0 & 24.4 & 44.6 & & 84.7 & 92.3 & 43.1 & 18.3 & 45.5 & 27.9 & & 59.4 & 78.7 & 67.6 & 48.9 \\
        BLIP-FF & 79.7 & 23.4 & 26.1 & 33.0 & 29.2 & 52.2 & & 80.0 & 89.9 & 22.8 & 28.9 & 41.0 & 22.4 & & 50.9 & 79.8 & 55.8 & 33.0 \\
        U-MARVEL & \textbf{84.4} & \textbf{47.3} & \textbf{33.6} & \textbf{34.2} & \textbf{36.4} & \textbf{60.7} & & \textbf{97.1} & \textbf{93.5} & \textbf{47.3} & \textbf{35.1} & \textbf{62.5} & \textbf{58.3} & & \textbf{78.8} & \textbf{88.5} & \textbf{79.4} & \textbf{74.7} \\
        \midrule
        
        \multicolumn{19}{c}{\textit{Generative Retrieval}} \\
        \midrule
        IRGen & 50.7 & -- & -- & -- & -- & -- & & -- & -- & -- & -- & -- & -- & & -- & -- & -- & -- \\
        GRACE & 39.5 & -- & -- & -- & -- & -- & & -- & -- & -- & -- & -- & -- & & -- & -- & -- & -- \\
        Baseline & 75.5 & 22.2 & 13.3 & 29.0 & 16.3 & 37.2 & & 38.4 & 89.3 & 22.2 & 14.9 & 41.5 & 19.7 & & 40.0 & 53.3 & 52.1 & 30.8 \\
        GENIUS$^{\mathcal{R}}$ & 78.0 & 27.4 & 16.2 & 30.2 & 19.3 & 39.5 & & 44.6 & \textbf{91.1} & 28.4& 16.3 & 41.9 & 20.7 & & 44.3 & 60.6 & 52.5 & 30.1 \\
        
        \rowcolor{gray!10}
        \textbf{WIDE (Ours)} & \textbf{79.8} & \textbf{30.2} & \textbf{17.6} & \textbf{31.5} & \textbf{21.7} & \textbf{41.5} & & \textbf{46.2} & 89.8 & \textbf{30.2} & \textbf{18.1} & \textbf{42.7} & \textbf{21.4} & & \textbf{44.8} & \textbf{63.1} & \textbf{53.2} & \textbf{32.4} \\
        \bottomrule
    \end{tabular}
    }
\end{table*}

Table~\ref{tab:overall_performance} presents the comprehensive evaluation of our WIDE framework against both embedding-based and generative baselines across the M-BEIR benchmark. The results yield several key observations regarding the effectiveness of uncertainty-guided wildcard decoding and hybrid re-ranking mechanisms.

WIDE achieves state-of-the-art performance across almost all tasks within the generative retrieval track, demonstrating significant margins over the strongest model, GENIUS$^{\mathcal{R}}$. Notably, in complex reasoning scenarios that exhibit severe cross-modal asymmetry, such as fine-grained visual manipulation (CIRR) and knowledge-based visual question retrieval (WebQA $q_t \rightarrow c_t$), WIDE outperforms GENIUS by 2.0 and 1.6 percentage points, respectively. Similar substantial gains are observed in image retrieval tasks like VisualNews (+2.8 percentage points). By deploying the AWD module to dynamically emit wildcards and the BSR module to re-rank candidates affected by identifier hallucinations, WIDE successfully avoids the premature pruning of relevant retrieval paths in standard trie-constrained beam search.

Generative retrieval often underperforms embedding-based retrieval because continuous representations must be compressed into discrete identifiers, inevitably introducing quantization loss and reducing representational capacity. As shown in Table~\ref{tab:overall_performance}, WIDE reduces this performance gap. Without relying on exhaustive similarity computation over the entire database, our framework achieves comparable performance to BLIP-FF and CLIP-SF~\cite{wei2024uniir} on specific benchmarks, including NIGHTS and EDIS. This demonstrates that adaptive uncertainty handling can improve the effectiveness of generative retrieval while preserving the structural advantages of a unified prefix trie.

We also include U-MARVEL (built upon the Qwen2-VL-7B backbone)~\cite{li2026u} as a strong reference for multimodal retrieval. While U-MARVEL achieves superior performance on the benchmark, it relies on a large multimodal foundation model with billions of parameters, introducing a much larger computational footprint during inference. In contrast, WIDE is built upon a lightweight T5-small decoder. The competitive performance of WIDE against such a heavyweight model highlights the effectiveness of uncertainty-aware decoding as an efficient alternative to simply scaling model size.

\subsection{Ablation Study}
\label{sec:ablation}

\begin{table}[t]
    \centering
    \caption{Ablation Study of the WIDE Framework. Performance is evaluated by incrementally integrating wildcard decoding, adaptive thresholding, and re-ranking modules. The ``Fixed'' variant uses an empirically tuned static entropy threshold.}
    \label{tab:ablation}

    \resizebox{\columnwidth}{!}{
    \begin{tabular}{l ccc cccc}
        \toprule
        \multirow{2}{*}{\textbf{Model Variants}}
        & \multicolumn{3}{c}{\textbf{Components}}
        & COCO & VN & WebQA & FIQ \\
        \cmidrule(lr){2-4}
        & AWD & AET & BSR
        & \scriptsize{$q_t \rightarrow c_i$}
        & \scriptsize{$q_t \rightarrow c_i$}
        & \scriptsize{$q_t \rightarrow c_t$}
        & \scriptsize{$(q_i,q_t) \rightarrow c_i$} \\
        \midrule

        Baseline
        & & & &
        75.5 & 22.2 & 38.4 & 16.3 \\

        \midrule

        Row 1 (Fixed $\tau$)
        & \checkmark & & &
        76.9 & 24.8 & 41.0 & 18.1 \\

        Row 2 (Dynamic $\tau$)
        & \checkmark & \checkmark & &
        77.2 & 26.7 & 43.0 & 18.9 \\

        \rowcolor{gray!10}
        \textbf{WIDE (Full)}
        & \checkmark & \checkmark & \checkmark &
        \textbf{79.8} & \textbf{30.2} & \textbf{46.2} & \textbf{21.7} \\

        \bottomrule
    \end{tabular}
    }
\end{table}
To validate the individual contributions of the proposed components within the WIDE framework, we conduct a systematic ablation study. Table~\ref{tab:ablation} reports the performance variations on four representative datasets covering diverse cross-modal retrieval settings, including MS-COCO, VisualNews, WebQA, and FashionIQ. The baseline follows the standard generative retrieval pipeline with trie-constrained beam search and does not apply any uncertainty intervention.

\subsubsection{Effectiveness of Wildcard Decoding}
Row 1 of Table~\ref{tab:ablation} evaluates the effect of wildcard decoding using a fixed entropy threshold. Compared with the baseline, introducing wildcard decoding consistently improves retrieval performance across all four datasets, with gains ranging from 1.4 to 2.6 percentage points. These results show that replacing uncertain intermediate identifier generations with wildcards can retain relevant retrieval paths that would otherwise be prematurely pruned during trie-constrained decoding.

\subsubsection{Adaptive versus Fixed Thresholding}
Row 2 replaces the fixed entropy threshold with the proposed adaptive entropy thresholding module. Compared with the fixed-threshold variant, adaptive thresholding further improves performance on all four datasets, with gains of 0.3, 1.9, 2.0, and 0.8 percentage points on MS-COCO, VisualNews, WebQA, and FashionIQ, respectively. This consistent improvement indicates that a level-specific entropy criterion is more effective than a single fixed threshold for determining when wildcard decoding should be activated. By assigning different entropy thresholds to different RQ levels, AET more accurately identifies intermediate query semantic blind spots.

\subsubsection{Effectiveness of BSR}
The full WIDE framework further introduces BSR to re-rank the candidate pool produced by AWD and AET. Adding BSR improves performance by 2.6, 3.5, 3.2, and 2.8 percentage points on MS-COCO, VisualNews, WebQA, and FashionIQ, respectively. After wildcard decoding, candidates associated with the same beam satisfy the same reliable identifier constraints. BSR therefore combines generation confidence from the constrained levels with continuous embedding similarity to distinguish these candidates more effectively. The consistent gains over Row 2 demonstrate the importance of re-ranking the expanded candidate pool after wildcard decoding.

\subsection{Analysis}
\subsubsection{Precision of Adaptive Uncertainty Intervention}

\begin{figure}[t]
    \centering
    \includegraphics[width=\linewidth]{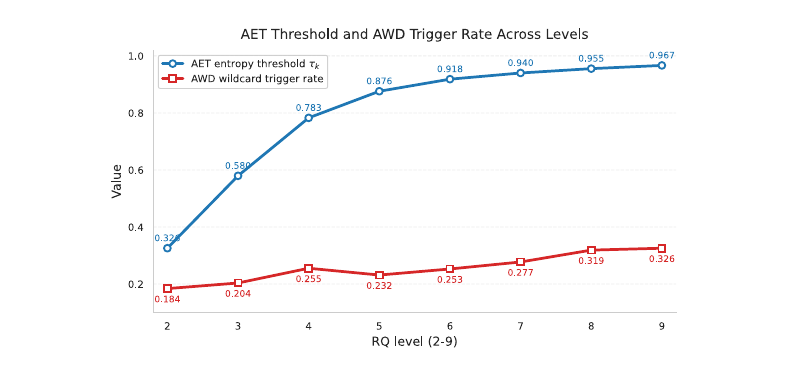}
    \caption{Layer-wise dynamics of the adaptive entropy threshold and wildcard trigger ratio.}
    \Description{A line chart over residual quantization levels showing that the adaptive entropy threshold rises toward deeper levels while the wildcard trigger ratio increases gradually.}
    \label{fig:layerwise_entropy_wildcard_dynamics}
\end{figure}

Figure \ref{fig:layerwise_entropy_wildcard_dynamics} illustrates the distribution of the dynamic entropy threshold alongside the corresponding wildcard trigger ratio across residual levels. A significant discrepancy in entropy exists between the shallow and deep quantization layers. Because the adaptive threshold derives from historical entropy statistics, this distinct upward trajectory demonstrates that the model inherently experiences greater uncertainty when generating identifiers at deeper levels compared to shallow ones. Shallow levels capture semantics that align well with text queries, resulting in low historical uncertainty. Deeper levels encode details typically absent from the text, which elevates the expected entropy boundary.

Despite the substantial increase in the entropy threshold at deeper levels, the wildcard trigger ratio exhibits a smooth and steady rise. This observation aligns with our fundamental design objective that wildcard activation should not concentrate disproportionately within specific architectural layers. Instead, the intervention relies entirely on the dynamic gap between the real-time predictive entropy and AET threshold. The steady trajectory of the trigger rate confirms that the AET module effectively regulates the decoding process. It prevents the system from emitting wildcards prematurely while successfully ensuring that no individual layer undergoes excessive candidate expansion.

\subsubsection{Candidate Expansion under Wildcard Decoding}

\begin{table}[t]
    \centering
    \caption{Wildcard Expansion Statistics of the WIDE Framework. The table reports the database size $|\mathcal{C}|$, the average number of wildcarded levels, and the candidate-set size induced by wildcard decoding before cross-beam aggregation and BSR re-ranking.}
    \label{tab:pool_size}
    \resizebox{\columnwidth}{!}{
    \begin{tabular}{l r c rrc}
        \toprule
        \multirow{2}{*}{\textbf{Dataset}}
        & \multicolumn{1}{c}{\multirow{2}{*}{\textbf{Database Size} $|\mathcal{C}|$}}
        & \textbf{Wildcarded Levels}
        & \multicolumn{3}{c}{\textbf{Wildcard Expansion Size}} \\
        \cmidrule(lr){3-3}
        \cmidrule(lr){4-6}
        & & Avg. $|\mathcal{A}_i|$
        & \multicolumn{1}{c}{Average}
        & \multicolumn{1}{c}{$P_{95}$}
        & Reduction (\%) \\
        \midrule
        MS-COCO     & 24,809    & 1.77 & 1.1     & 1.0   & 99.99\% \\
        Fashion200K & 201,824   & 1.96 & 21.9    & 34.0  & 99.99\% \\
        WebQA       & 544,457   & 1.90 & 276.8   & 2.0   & 99.95\% \\
        CIRR        & 21,551    & 1.92 & 55.0    & 3.0   & 99.74\% \\
        VisualNews  & 542,246   & 1.93 & 1,654.2 & 57.0  & 99.69\% \\
        EDIS        & 1,047,067 & 2.06 & 2,357.7 & 224.0 & 99.77\% \\
        \midrule
        \rowcolor{gray!10}
        \textbf{Query-weighted Avg.}
        & \textbf{--}
        & \textbf{1.97}
        & \textbf{1,142.7}
        & \textbf{--}
        & \textbf{99.76\%} \\
        \bottomrule
    \end{tabular}
    }
\end{table}

Table~\ref{tab:pool_size} analyzes the candidate expansion induced by wildcard decoding relative to the full database size $|\mathcal{C}|$. Rather than measuring the final candidate pool after aggregating all source beams, the reported statistics characterize the candidates matched by wildcard-based expansion while preserving the remaining identifier constraints. Across the evaluated datasets, this local expansion retains only a small fraction of the database, with a query-weighted reduction of 99.76\%.

On average, wildcard decoding is activated at 1.97 identifier levels. This relatively small number indicates that AWD modifies only a limited portion of the identifier sequence, while the remaining levels continue to constrain candidate matching. Consequently, uncertain intermediate generations can be relaxed without removing the discrete constraints provided by the rest of the identifier.

The expansion size also exhibits a strongly skewed distribution on several datasets. For example, WebQA has an average expansion size of 276.8 but a $P_{95}$ of only 2.0, indicating that most wildcard activations correspond to very small candidate sets, while a small number of wildcard patterns match substantially more candidates and increase the average. Similar behavior can be observed on CIRR, VisualNews, and EDIS.
\subsubsection{Qualitative Analysis of Retrieval Results}

\begin{figure}[t]
    \centering
    \includegraphics[width=0.95\linewidth]{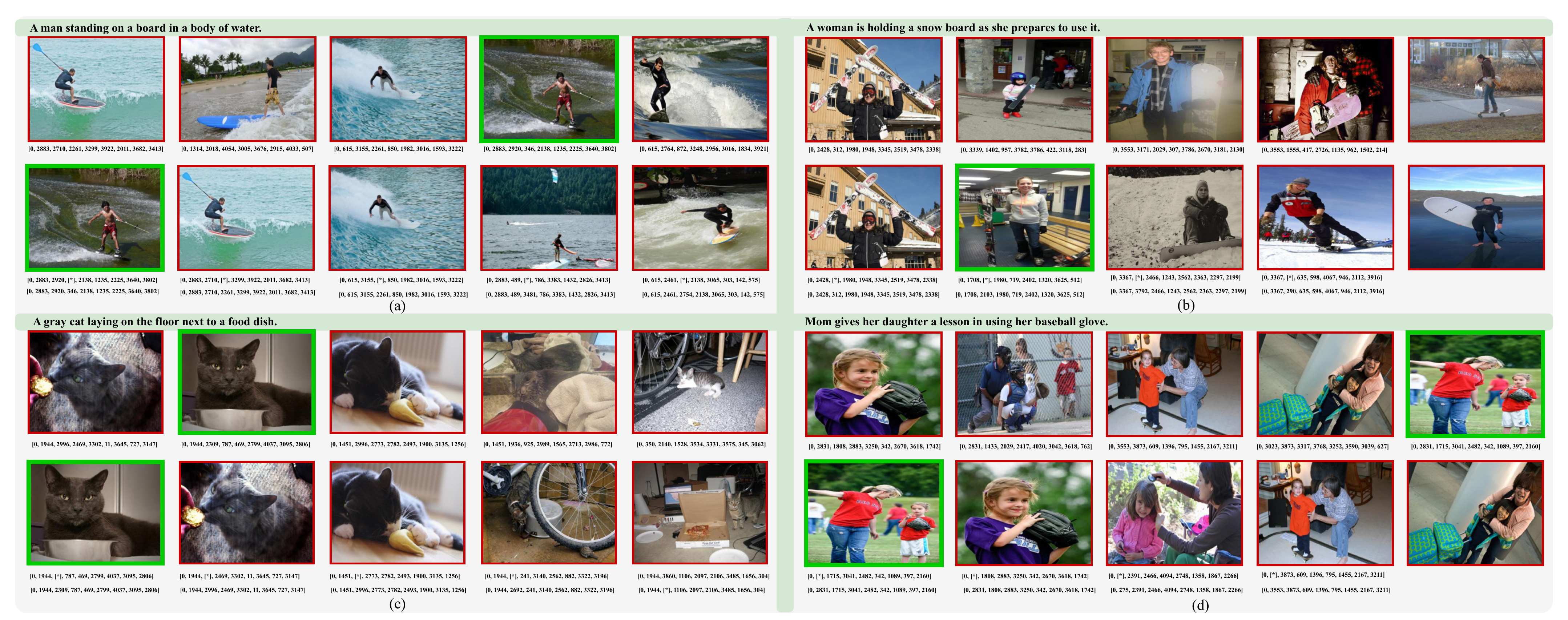}
    \caption{Qualitative comparison of retrieval results. For each query, the first and second rows show the top-5 results retrieved by the baseline and WIDE, respectively. The numbers below each candidate denote its discrete identifier, and the wildcarded identifier shown for WIDE marks the levels replaced by ``*'' during wildcard decoding.}
    \Description{Four qualitative retrieval examples comparing the baseline and WIDE top-five results. Each candidate is annotated with its discrete identifier, and asterisks mark identifier levels replaced by wildcards in WIDE.}
    \label{fig:qualitative_retrieval_comparison}
\end{figure}

To comprehensively illustrate the advantages of the WIDE framework over the baseline, we present qualitative retrieval results in Figure~\ref{fig:qualitative_retrieval_comparison}. Specifically, Figure~\ref{fig:qualitative_retrieval_comparison}(a) demonstrates that the framework effectively resolves cross-modal information asymmetry by utilizing the BSR module to compute hybrid scores, which elevates the rank of true targets containing rich visual details absent from concise text queries. Furthermore, Figure~\ref{fig:qualitative_retrieval_comparison}(b) illustrates robust candidate pool recovery. By substituting deterministic identifier generation at highly uncertain levels with wildcards, WIDE successfully retrieves ground truths that the baseline entirely misses or ranks exceptionally low. As evidenced in Figure~\ref{fig:qualitative_retrieval_comparison}(c), even when the baseline already positions the target near the top of the retrieved list, the framework further refines the ranking to achieve the highest possible placement. Finally, Figure~\ref{fig:qualitative_retrieval_comparison}(d) highlights a strong capacity for distractor mitigation in complex scenes, where WIDE consistently prioritizes the correct target over visually similar but inaccurate candidates.

\section{Limitations}
\label{sec:limitations}

WIDE improves retrieval accuracy by expanding candidate paths at uncertain identifier levels, which inevitably introduces additional decoding and re-ranking overhead compared with standard trie-constrained beam search despite the restricted wildcard activation. The expanded candidate pool is further re-ranked using continuous embedding similarity, meaning that the final retrieval stage still partially relies on the semantic discrimination provided by embedding-based representations. Moreover, WIDE relies on entropy deviations to identify query semantic blind spots, while high uncertainty may also arise from noisy or ambiguous queries or insufficient decoder confidence, potentially causing unnecessary wildcard activation.

\section{Conclusion}
\label{sec:conclusion}

In this paper, we revisit cross-modal generative retrieval from the perspective of information asymmetry and identify forced hallucination as an important failure mode in discrete identifier generation. To address this issue, we propose WIDE, which introduces wildcard inference with dynamic expansion to relax unreliable decoding decisions and preserve alternative retrieval paths. WIDE further incorporates a hybrid re-ranking strategy that combines reliable generative evidence with continuous semantic similarity to distinguish the expanded candidates. Experiments on the M-BEIR benchmark show that WIDE provides clear improvements over existing generative retrieval methods across diverse cross-modal retrieval tasks. Overall, our results demonstrate the effectiveness of explicitly handling uncertainty in cross-modal generative retrieval while retaining its discrete retrieval formulation.

\begin{acks}
This work was supported by the Department of Science and Technology of Jilin Province (Grant Number: 20260301004YY).
\end{acks}


\bibliographystyle{ACM-Reference-Format}
\bibliography{main}


\begin{thebibliography}{41}


\ifx \showCODEN    \undefined \def \showCODEN     #1{\unskip}     \fi
\ifx \showISBNx    \undefined \def \showISBNx     #1{\unskip}     \fi
\ifx \showISBNxiii \undefined \def \showISBNxiii  #1{\unskip}     \fi
\ifx \showISSN     \undefined \def \showISSN      #1{\unskip}     \fi
\ifx \showLCCN     \undefined \def \showLCCN      #1{\unskip}     \fi
\ifx \shownote     \undefined \def \shownote      #1{#1}          \fi
\ifx \showarticletitle \undefined \def \showarticletitle #1{#1}   \fi
\ifx \showURL      \undefined \def \showURL       {\relax}        \fi
\providecommand\bibfield[2]{#2}
\providecommand\bibinfo[2]{#2}
\providecommand\natexlab[1]{#1}
\providecommand\showeprint[2][]{arXiv:#2}

\bibitem[Anderson et~al\mbox{.}(2017)]%
        {anderson2017guided}
\bibfield{author}{\bibinfo{person}{Peter Anderson}, \bibinfo{person}{Basura Fernando}, \bibinfo{person}{Mark Johnson}, {and} \bibinfo{person}{Stephen Gould}.} \bibinfo{year}{2017}\natexlab{}.
\newblock \showarticletitle{Guided Open Vocabulary Image Captioning with Constrained Beam Search}. In \bibinfo{booktitle}{\emph{Proceedings of the 2017 Conference on Empirical Methods in Natural Language Processing, {EMNLP} 2017, Copenhagen, Denmark, September 9-11, 2017}}. \bibinfo{publisher}{Association for Computational Linguistics}, \bibinfo{pages}{936--945}.
\newblock


\bibitem[Cao et~al\mbox{.}(2021)]%
        {cao2020genre}
\bibfield{author}{\bibinfo{person}{Nicola~De Cao}, \bibinfo{person}{Gautier Izacard}, \bibinfo{person}{Sebastian Riedel}, {and} \bibinfo{person}{Fabio Petroni}.} \bibinfo{year}{2021}\natexlab{}.
\newblock \showarticletitle{Autoregressive Entity Retrieval}. In \bibinfo{booktitle}{\emph{9th International Conference on Learning Representations, {ICLR} 2021, Virtual Event, Austria, May 3-7, 2021}}. \bibinfo{publisher}{OpenReview.net}.
\newblock


\bibitem[Chang et~al\mbox{.}(2022)]%
        {chang2022webqa}
\bibfield{author}{\bibinfo{person}{Yingshan Chang}, \bibinfo{person}{Mridu Narang}, \bibinfo{person}{Hisami Suzuki}, \bibinfo{person}{Guihong Cao}, \bibinfo{person}{Jianfeng Gao}, {and} \bibinfo{person}{Yonatan Bisk}.} \bibinfo{year}{2022}\natexlab{}.
\newblock \showarticletitle{Webqa: Multihop and multimodal qa}. In \bibinfo{booktitle}{\emph{Proceedings of the IEEE/CVF conference on computer vision and pattern recognition}}. \bibinfo{pages}{16495--16504}.
\newblock


\bibitem[Chen et~al\mbox{.}(2022)]%
        {chen2022corpusbrain}
\bibfield{author}{\bibinfo{person}{Jiangui Chen}, \bibinfo{person}{Ruqing Zhang}, \bibinfo{person}{Jiafeng Guo}, \bibinfo{person}{Yiqun Liu}, \bibinfo{person}{Yixing Fan}, {and} \bibinfo{person}{Xueqi Cheng}.} \bibinfo{year}{2022}\natexlab{}.
\newblock \showarticletitle{CorpusBrain: Pre-train a Generative Retrieval Model for Knowledge-Intensive Language Tasks}. In \bibinfo{booktitle}{\emph{Proceedings of the 31st {ACM} International Conference on Information {\&} Knowledge Management, Atlanta, GA, USA, October 17-21, 2022}}. \bibinfo{publisher}{{ACM}}, \bibinfo{pages}{191--200}.
\newblock


\bibitem[Chen et~al\mbox{.}(2023)]%
        {chen2023can}
\bibfield{author}{\bibinfo{person}{Yang Chen}, \bibinfo{person}{Hexiang Hu}, \bibinfo{person}{Yi Luan}, \bibinfo{person}{Haitian Sun}, \bibinfo{person}{Soravit Changpinyo}, \bibinfo{person}{Alan Ritter}, {and} \bibinfo{person}{Ming-Wei Chang}.} \bibinfo{year}{2023}\natexlab{}.
\newblock \showarticletitle{Can pre-trained vision and language models answer visual information-seeking questions?}. In \bibinfo{booktitle}{\emph{Proceedings of the 2023 Conference on Empirical Methods in Natural Language Processing}}. \bibinfo{pages}{14948--14968}.
\newblock


\bibitem[Fang et~al\mbox{.}(2025)]%
        {fang2025cart}
\bibfield{author}{\bibinfo{person}{Minghui Fang}, \bibinfo{person}{Shengpeng Ji}, \bibinfo{person}{Jialong Zuo}, \bibinfo{person}{Hai Huang}, \bibinfo{person}{Yan Xia}, \bibinfo{person}{Jieming Zhu}, \bibinfo{person}{Xize Cheng}, \bibinfo{person}{Xiaoda Yang}, \bibinfo{person}{Wenrui Liu}, \bibinfo{person}{Gang Wang}, \bibinfo{person}{Zhenhua Dong}, {and} \bibinfo{person}{Zhou Zhao}.} \bibinfo{year}{2025}\natexlab{}.
\newblock \showarticletitle{{CART:} {A} Generative Cross-Modal Retrieval Framework With Coarse-To-Fine Semantic Modeling}. In \bibinfo{booktitle}{\emph{Proceedings of the 63rd Annual Meeting of the Association for Computational Linguistics (Volume 1: Long Papers), {ACL} 2025, Vienna, Austria, July 27 - August 1, 2025}}. \bibinfo{publisher}{Association for Computational Linguistics}, \bibinfo{pages}{15120--15133}.
\newblock


\bibitem[Fu et~al\mbox{.}(2023)]%
        {fu2023dreamsim}
\bibfield{author}{\bibinfo{person}{Stephanie Fu}, \bibinfo{person}{Netanel Tamir}, \bibinfo{person}{Shobhita Sundaram}, \bibinfo{person}{Lucy Chai}, \bibinfo{person}{Richard Zhang}, \bibinfo{person}{Tali Dekel}, {and} \bibinfo{person}{Phillip Isola}.} \bibinfo{year}{2023}\natexlab{}.
\newblock \showarticletitle{DreamSim: Learning new dimensions of human visual similarity using synthetic data}. In \bibinfo{booktitle}{\emph{Advances in Neural Information Processing Systems}}, Vol.~\bibinfo{volume}{36}.
\newblock


\bibitem[Han et~al\mbox{.}(2017)]%
        {han2017automatic}
\bibfield{author}{\bibinfo{person}{Xintong Han}, \bibinfo{person}{Zuxuan Wu}, \bibinfo{person}{Phoenix~X Huang}, \bibinfo{person}{Xiao Zhang}, \bibinfo{person}{Menglong Zhu}, \bibinfo{person}{Yuan Li}, \bibinfo{person}{Yang Zhao}, {and} \bibinfo{person}{Larry~S Davis}.} \bibinfo{year}{2017}\natexlab{}.
\newblock \showarticletitle{Automatic spatially-aware fashion concept discovery}. In \bibinfo{booktitle}{\emph{Proceedings of the IEEE international conference on computer vision}}. \bibinfo{pages}{1202--1211}.
\newblock


\bibitem[Hendrycks and Gimpel(2017)]%
        {hendrycks2017baseline}
\bibfield{author}{\bibinfo{person}{Dan Hendrycks} {and} \bibinfo{person}{Kevin Gimpel}.} \bibinfo{year}{2017}\natexlab{}.
\newblock \showarticletitle{A Baseline for Detecting Misclassified and Out-of-Distribution Examples in Neural Networks}. In \bibinfo{booktitle}{\emph{International Conference on Learning Representations}}.
\newblock


\bibitem[Hokamp and Liu(2017)]%
        {hokamp2017lexically}
\bibfield{author}{\bibinfo{person}{Chris Hokamp} {and} \bibinfo{person}{Qun Liu}.} \bibinfo{year}{2017}\natexlab{}.
\newblock \showarticletitle{Lexically Constrained Decoding for Sequence Generation Using Grid Beam Search}. In \bibinfo{booktitle}{\emph{Proceedings of the 55th Annual Meeting of the Association for Computational Linguistics, {ACL} 2017, Vancouver, Canada, July 30 - August 4, Volume 1: Long Papers}}. \bibinfo{publisher}{Association for Computational Linguistics}, \bibinfo{pages}{1535--1546}.
\newblock


\bibitem[Holtzman et~al\mbox{.}(2020)]%
        {holtzman2019curious}
\bibfield{author}{\bibinfo{person}{Ari Holtzman}, \bibinfo{person}{Jan Buys}, \bibinfo{person}{Li Du}, \bibinfo{person}{Maxwell Forbes}, {and} \bibinfo{person}{Yejin Choi}.} \bibinfo{year}{2020}\natexlab{}.
\newblock \showarticletitle{The Curious Case of Neural Text Degeneration}. In \bibinfo{booktitle}{\emph{International Conference on Learning Representations}}.
\newblock


\bibitem[Hu et~al\mbox{.}(2023)]%
        {hu2023oven}
\bibfield{author}{\bibinfo{person}{Hexiang Hu}, \bibinfo{person}{Yi Luan}, \bibinfo{person}{Yang Chen}, \bibinfo{person}{Urvashi Khandelwal}, \bibinfo{person}{Mandar Joshi}, \bibinfo{person}{Kenton Lee}, \bibinfo{person}{Kristina Toutanova}, {and} \bibinfo{person}{Ming-Wei Chang}.} \bibinfo{year}{2023}\natexlab{}.
\newblock \showarticletitle{Open-domain visual entity recognition: Towards recognizing millions of wikipedia entities}. In \bibinfo{booktitle}{\emph{Proceedings of the IEEE/CVF International Conference on Computer Vision}}. \bibinfo{pages}{11206--11216}.
\newblock


\bibitem[Kim et~al\mbox{.}(2025)]%
        {kim2025genius}
\bibfield{author}{\bibinfo{person}{Sungyeon Kim}, \bibinfo{person}{Xinliang Zhu}, \bibinfo{person}{Xiaofan Lin}, \bibinfo{person}{Muhammet Bastan}, \bibinfo{person}{Douglas Gray}, {and} \bibinfo{person}{Suha Kwak}.} \bibinfo{year}{2025}\natexlab{}.
\newblock \showarticletitle{GENIUS: A generative framework for universal multimodal search}. In \bibinfo{booktitle}{\emph{Proceedings of the Computer Vision and Pattern Recognition Conference}}. \bibinfo{pages}{19659--19669}.
\newblock


\bibitem[Kuhn et~al\mbox{.}(2023)]%
        {kuhn2023semantic}
\bibfield{author}{\bibinfo{person}{Lorenz Kuhn}, \bibinfo{person}{Yarin Gal}, {and} \bibinfo{person}{Sebastian Farquhar}.} \bibinfo{year}{2023}\natexlab{}.
\newblock \showarticletitle{Semantic Uncertainty: Linguistic Invariances for Uncertainty Estimation in Natural Language Generation}. In \bibinfo{booktitle}{\emph{International Conference on Learning Representations}}.
\newblock


\bibitem[Lee et~al\mbox{.}(2022)]%
        {lee2022autoregressive}
\bibfield{author}{\bibinfo{person}{Doyup Lee}, \bibinfo{person}{Chiheon Kim}, \bibinfo{person}{Saehoon Kim}, \bibinfo{person}{Minsu Cho}, {and} \bibinfo{person}{Wook-Shin Han}.} \bibinfo{year}{2022}\natexlab{}.
\newblock \showarticletitle{Autoregressive Image Generation using Residual Quantization}. In \bibinfo{booktitle}{\emph{Proceedings of the IEEE/CVF Conference on Computer Vision and Pattern Recognition}}. \bibinfo{pages}{11523--11532}.
\newblock


\bibitem[Li et~al\mbox{.}(2025a)]%
        {li2025semcore}
\bibfield{author}{\bibinfo{person}{Haoxuan Li}, \bibinfo{person}{Yi Bin}, \bibinfo{person}{Yunshan Ma}, \bibinfo{person}{Guoqing Wang}, \bibinfo{person}{Yang Yang}, \bibinfo{person}{See{-}Kiong Ng}, {and} \bibinfo{person}{Tat{-}Seng Chua}.} \bibinfo{year}{2025}\natexlab{a}.
\newblock \showarticletitle{SemCORE: {A} Semantic-Enhanced Generative Cross-Modal Retrieval Framework with MLLMs}.
\newblock \bibinfo{journal}{\emph{CoRR}}  \bibinfo{volume}{abs/2504.13172} (\bibinfo{year}{2025}).
\newblock


\bibitem[Li et~al\mbox{.}(2025c)]%
        {li2025matching}
\bibfield{author}{\bibinfo{person}{Xiaoxi Li}, \bibinfo{person}{Jiajie Jin}, \bibinfo{person}{Yujia Zhou}, \bibinfo{person}{Yuyao Zhang}, \bibinfo{person}{Peitian Zhang}, \bibinfo{person}{Yutao Zhu}, {and} \bibinfo{person}{Zhicheng Dou}.} \bibinfo{year}{2025}\natexlab{c}.
\newblock \showarticletitle{From matching to generation: A survey on generative information retrieval}.
\newblock \bibinfo{journal}{\emph{ACM Transactions on Information Systems}} \bibinfo{volume}{43}, \bibinfo{number}{3} (\bibinfo{year}{2025}), \bibinfo{pages}{1--62}.
\newblock


\bibitem[Li et~al\mbox{.}(2026)]%
        {li2026u}
\bibfield{author}{\bibinfo{person}{Xiaojie Li}, \bibinfo{person}{Chu Li}, \bibinfo{person}{Shi-Zhe Chen}, {and} \bibinfo{person}{Xi Chen}.} \bibinfo{year}{2026}\natexlab{}.
\newblock \showarticletitle{U-MARVEL: Unveiling Key Factors for Universal Multimodal Retrieval via Embedding Learning with MLLMs}. In \bibinfo{booktitle}{\emph{International Conference on Learning Representations}}.
\newblock


\bibitem[Li et~al\mbox{.}(2025b)]%
        {li2025revolutionizing}
\bibfield{author}{\bibinfo{person}{Yongqi Li}, \bibinfo{person}{Hongru Cai}, \bibinfo{person}{Wenjie Wang}, \bibinfo{person}{Leigang Qu}, \bibinfo{person}{Yinwei Wei}, \bibinfo{person}{Wenjie Li}, \bibinfo{person}{Liqiang Nie}, {and} \bibinfo{person}{Tat{-}Seng Chua}.} \bibinfo{year}{2025}\natexlab{b}.
\newblock \showarticletitle{Revolutionizing Text-to-Image Retrieval as Autoregressive Token-to-Voken Generation}. In \bibinfo{booktitle}{\emph{Proceedings of the 48th International {ACM} {SIGIR} Conference on Research and Development in Information Retrieval, {SIGIR} 2025, Padua, Italy, July 13-18, 2025}}. \bibinfo{publisher}{{ACM}}, \bibinfo{pages}{813--822}.
\newblock


\bibitem[Li et~al\mbox{.}(2024)]%
        {li2024generative}
\bibfield{author}{\bibinfo{person}{Yongqi Li}, \bibinfo{person}{Wenjie Wang}, \bibinfo{person}{Leigang Qu}, \bibinfo{person}{Liqiang Nie}, \bibinfo{person}{Wenjie Li}, {and} \bibinfo{person}{Tat{-}Seng Chua}.} \bibinfo{year}{2024}\natexlab{}.
\newblock \showarticletitle{Generative Cross-Modal Retrieval: Memorizing Images in Multimodal Language Models for Retrieval and Beyond}. In \bibinfo{booktitle}{\emph{Proceedings of the 62nd Annual Meeting of the Association for Computational Linguistics (Volume 1: Long Papers), {ACL} 2024, Bangkok, Thailand, August 11-16, 2024}}, \bibfield{editor}{\bibinfo{person}{Lun{-}Wei Ku}, \bibinfo{person}{Andre Martins}, {and} \bibinfo{person}{Vivek Srikumar}} (Eds.). \bibinfo{publisher}{Association for Computational Linguistics}, \bibinfo{pages}{11851--11861}.
\newblock


\bibitem[Liang et~al\mbox{.}(2022)]%
        {liang2022mind}
\bibfield{author}{\bibinfo{person}{Weixin Liang}, \bibinfo{person}{Yuhui Zhang}, \bibinfo{person}{Yongchan Kwon}, \bibinfo{person}{Serena Yeung}, {and} \bibinfo{person}{James Zou}.} \bibinfo{year}{2022}\natexlab{}.
\newblock \showarticletitle{Mind the Gap: Understanding the Modality Gap in Multi-modal Contrastive Representation Learning}. In \bibinfo{booktitle}{\emph{Advances in Neural Information Processing Systems 35: Annual Conference on Neural Information Processing Systems 2022, NeurIPS 2022, New Orleans, LA, USA, November 28 - December 9, 2022}}.
\newblock


\bibitem[Lin et~al\mbox{.}(2014)]%
        {lin2014microsoft}
\bibfield{author}{\bibinfo{person}{Tsung-Yi Lin}, \bibinfo{person}{Michael Maire}, \bibinfo{person}{Serge Belongie}, \bibinfo{person}{James Hays}, \bibinfo{person}{Pietro Perona}, \bibinfo{person}{Deva Ramanan}, \bibinfo{person}{Piotr Doll{\'a}r}, {and} \bibinfo{person}{C~Lawrence Zitnick}.} \bibinfo{year}{2014}\natexlab{}.
\newblock \showarticletitle{Microsoft COCO: Common objects in context}. In \bibinfo{booktitle}{\emph{European conference on computer vision}}. Springer, \bibinfo{pages}{740--755}.
\newblock


\bibitem[Liu et~al\mbox{.}(2021b)]%
        {liu2021visualnews}
\bibfield{author}{\bibinfo{person}{Fuxiao Liu}, \bibinfo{person}{Yinghan Wang}, \bibinfo{person}{Tianjun Wang}, {and} \bibinfo{person}{Vicente Ordonez}.} \bibinfo{year}{2021}\natexlab{b}.
\newblock \showarticletitle{VisualNews: Benchmark and Challenges in News Image Captioning}. In \bibinfo{booktitle}{\emph{Proceedings of the 2021 Conference on Empirical Methods in Natural Language Processing}}. \bibinfo{pages}{6761--6771}.
\newblock


\bibitem[Liu et~al\mbox{.}(2023)]%
        {liu2023edis}
\bibfield{author}{\bibinfo{person}{Siqi Liu}, \bibinfo{person}{Weixi Feng}, \bibinfo{person}{Tsu-Jui Fu}, \bibinfo{person}{Wenhu Chen}, {and} \bibinfo{person}{William Wang}.} \bibinfo{year}{2023}\natexlab{}.
\newblock \showarticletitle{EDIS: Entity-Driven Image Search over Multimodal Web Content}. In \bibinfo{booktitle}{\emph{Proceedings of the 2023 Conference on Empirical Methods in Natural Language Processing}}. \bibinfo{pages}{4877--4894}.
\newblock


\bibitem[Liu et~al\mbox{.}(2021a)]%
        {liu2021cirr}
\bibfield{author}{\bibinfo{person}{Zheyuan Liu}, \bibinfo{person}{Cristian Rodriguez-Opazo}, \bibinfo{person}{Damien Teney}, {and} \bibinfo{person}{Stephen Gould}.} \bibinfo{year}{2021}\natexlab{a}.
\newblock \showarticletitle{Image Retrieval on Real-life Images with Pre-trained Vision-and-Language Models}. In \bibinfo{booktitle}{\emph{Proceedings of the IEEE/CVF International Conference on Computer Vision}}. \bibinfo{pages}{2125--2134}.
\newblock


\bibitem[Loshchilov and Hutter(2019)]%
        {loshchilov2017decoupled}
\bibfield{author}{\bibinfo{person}{Ilya Loshchilov} {and} \bibinfo{person}{Frank Hutter}.} \bibinfo{year}{2019}\natexlab{}.
\newblock \showarticletitle{Decoupled Weight Decay Regularization}. In \bibinfo{booktitle}{\emph{International Conference on Learning Representations}}.
\newblock


\bibitem[Malinin and Gales(2021)]%
        {malinin2020uncertainty}
\bibfield{author}{\bibinfo{person}{Andrey Malinin} {and} \bibinfo{person}{Mark J.~F. Gales}.} \bibinfo{year}{2021}\natexlab{}.
\newblock \showarticletitle{Uncertainty Estimation in Autoregressive Structured Prediction}. In \bibinfo{booktitle}{\emph{9th International Conference on Learning Representations, {ICLR} 2021, Virtual Event, Austria, May 3-7, 2021}}. \bibinfo{publisher}{OpenReview.net}.
\newblock


\bibitem[Qu et~al\mbox{.}(2025)]%
        {qu2024tiger}
\bibfield{author}{\bibinfo{person}{Leigang Qu}, \bibinfo{person}{Haochuan Li}, \bibinfo{person}{Tan Wang}, \bibinfo{person}{Wenjie Wang}, \bibinfo{person}{Yongqi Li}, \bibinfo{person}{Liqiang Nie}, {and} \bibinfo{person}{Tat{-}Seng Chua}.} \bibinfo{year}{2025}\natexlab{}.
\newblock \showarticletitle{TIGeR: Unifying Text-to-Image Generation and Retrieval with Large Multimodal Models}. In \bibinfo{booktitle}{\emph{The Thirteenth International Conference on Learning Representations, {ICLR} 2025, Singapore, April 24-28, 2025}}. \bibinfo{publisher}{OpenReview.net}.
\newblock


\bibitem[Radford et~al\mbox{.}(2021)]%
        {radford2021learning}
\bibfield{author}{\bibinfo{person}{Alec Radford}, \bibinfo{person}{Jong~Wook Kim}, \bibinfo{person}{Chris Hallacy}, \bibinfo{person}{Aditya Ramesh}, \bibinfo{person}{Gabriel Goh}, \bibinfo{person}{Sandhini Agarwal}, \bibinfo{person}{Girish Sastry}, \bibinfo{person}{Amanda Askell}, \bibinfo{person}{Pamela Mishkin}, \bibinfo{person}{Jack Clark}, \bibinfo{person}{Gretchen Krueger}, {and} \bibinfo{person}{Ilya Sutskever}.} \bibinfo{year}{2021}\natexlab{}.
\newblock \showarticletitle{Learning transferable visual models from natural language supervision}. In \bibinfo{booktitle}{\emph{International Conference on Machine Learning}}. PMLR, \bibinfo{pages}{8748--8763}.
\newblock


\bibitem[Raffel et~al\mbox{.}(2020)]%
        {raffel2020exploring}
\bibfield{author}{\bibinfo{person}{Colin Raffel}, \bibinfo{person}{Noam Shazeer}, \bibinfo{person}{Adam Roberts}, \bibinfo{person}{Katherine Lee}, \bibinfo{person}{Sharan Narang}, \bibinfo{person}{Michael Matena}, \bibinfo{person}{Yanqi Zhou}, \bibinfo{person}{Wei Li}, {and} \bibinfo{person}{Peter~J Liu}.} \bibinfo{year}{2020}\natexlab{}.
\newblock \showarticletitle{Exploring the limits of transfer learning with a unified text-to-text transformer}.
\newblock \bibinfo{journal}{\emph{Journal of Machine Learning Research}} \bibinfo{volume}{21}, \bibinfo{number}{140} (\bibinfo{year}{2020}), \bibinfo{pages}{1--67}.
\newblock


\bibitem[Song et~al\mbox{.}(2024)]%
        {song2024re3val}
\bibfield{author}{\bibinfo{person}{EuiYul Song}, \bibinfo{person}{Sangryul Kim}, \bibinfo{person}{Haeju Lee}, \bibinfo{person}{Joonkee Kim}, {and} \bibinfo{person}{James Thorne}.} \bibinfo{year}{2024}\natexlab{}.
\newblock \showarticletitle{Re3val: Reinforced and Reranked Generative Retrieval}. In \bibinfo{booktitle}{\emph{Findings of the Association for Computational Linguistics: {EACL} 2024, St. Julian's, Malta, March 17-22, 2024}} \emph{(\bibinfo{series}{Findings of {ACL}})}. \bibinfo{publisher}{Association for Computational Linguistics}, \bibinfo{pages}{393--409}.
\newblock


\bibitem[Stahlberg and Byrne(2019)]%
        {stahlberg2019nmt}
\bibfield{author}{\bibinfo{person}{Felix Stahlberg} {and} \bibinfo{person}{Bill Byrne}.} \bibinfo{year}{2019}\natexlab{}.
\newblock \showarticletitle{On NMT Search Errors and Model Errors: Cat Got Your Tongue?}. In \bibinfo{booktitle}{\emph{Proceedings of the 2019 Conference on Empirical Methods in Natural Language Processing and the 9th International Joint Conference on Natural Language Processing (EMNLP-IJCNLP)}}. \bibinfo{pages}{3356--3362}.
\newblock


\bibitem[Tay et~al\mbox{.}(2022)]%
        {tay2022dsi}
\bibfield{author}{\bibinfo{person}{Yi Tay}, \bibinfo{person}{Vinh Tran}, \bibinfo{person}{Mostafa Dehghani}, \bibinfo{person}{Jianmo Ni}, \bibinfo{person}{Dara Bahri}, \bibinfo{person}{Harsh Mehta}, \bibinfo{person}{Zhen Qin}, \bibinfo{person}{Kai Hui}, \bibinfo{person}{Zhe Zhao}, \bibinfo{person}{Jai~Prakash Gupta}, \bibinfo{person}{Tal Schuster}, \bibinfo{person}{William~W. Cohen}, {and} \bibinfo{person}{Donald Metzler}.} \bibinfo{year}{2022}\natexlab{}.
\newblock \showarticletitle{Transformer Memory as a Differentiable Search Index}. In \bibinfo{booktitle}{\emph{Advances in Neural Information Processing Systems 35: Annual Conference on Neural Information Processing Systems 2022, NeurIPS 2022, New Orleans, LA, USA, November 28 - December 9, 2022}}, \bibfield{editor}{\bibinfo{person}{Sanmi Koyejo}, \bibinfo{person}{S.~Mohamed}, \bibinfo{person}{A.~Agarwal}, \bibinfo{person}{Danielle Belgrave}, \bibinfo{person}{K.~Cho}, {and} \bibinfo{person}{A.~Oh}} (Eds.).
\newblock


\bibitem[Wang et~al\mbox{.}(2022)]%
        {wang2022neural}
\bibfield{author}{\bibinfo{person}{Yujing Wang}, \bibinfo{person}{Yingyan Hou}, \bibinfo{person}{Haonan Wang}, \bibinfo{person}{Ziming Miao}, \bibinfo{person}{Shibin Wu}, \bibinfo{person}{Qi Chen}, \bibinfo{person}{Yuqing Xia}, \bibinfo{person}{Chengmin Chi}, \bibinfo{person}{Guoshuai Zhao}, \bibinfo{person}{Zheng Liu}, \bibinfo{person}{Xing Xie}, \bibinfo{person}{Hao Sun}, \bibinfo{person}{Weiwei Deng}, \bibinfo{person}{Qi Zhang}, {and} \bibinfo{person}{Mao Yang}.} \bibinfo{year}{2022}\natexlab{}.
\newblock \showarticletitle{A Neural Corpus Indexer for Document Retrieval}. In \bibinfo{booktitle}{\emph{Advances in Neural Information Processing Systems 35: Annual Conference on Neural Information Processing Systems 2022, NeurIPS 2022, New Orleans, LA, USA, November 28 - December 9, 2022}}.
\newblock


\bibitem[Wei et~al\mbox{.}(2024)]%
        {wei2024uniir}
\bibfield{author}{\bibinfo{person}{Cong Wei}, \bibinfo{person}{Yang Chen}, \bibinfo{person}{Haonan Chen}, \bibinfo{person}{Hexiang Hu}, \bibinfo{person}{Ge Zhang}, \bibinfo{person}{Jie Fu}, \bibinfo{person}{Alan Ritter}, {and} \bibinfo{person}{Wenhu Chen}.} \bibinfo{year}{2024}\natexlab{}.
\newblock \showarticletitle{UniIR: Training and Benchmarking Universal Multimodal Information Retrievers}. In \bibinfo{booktitle}{\emph{European Conference on Computer Vision}}. Springer.
\newblock


\bibitem[Wu et~al\mbox{.}(2021)]%
        {wu2021fashion}
\bibfield{author}{\bibinfo{person}{Hui Wu}, \bibinfo{person}{Yupeng Gao}, \bibinfo{person}{Xiaoxiao Guo}, \bibinfo{person}{Ziad Al-Halah}, \bibinfo{person}{Steven Rennie}, \bibinfo{person}{Kristen Grauman}, {and} \bibinfo{person}{Rogerio Feris}.} \bibinfo{year}{2021}\natexlab{}.
\newblock \showarticletitle{Fashion IQ: A new dataset towards retrieving images by natural language feedback}. In \bibinfo{booktitle}{\emph{Proceedings of the IEEE/CVF Conference on Computer Vision and Pattern Recognition}}. \bibinfo{pages}{11307--11317}.
\newblock


\bibitem[Xia et~al\mbox{.}(2025)]%
        {xia2025survey}
\bibfield{author}{\bibinfo{person}{Zhiqiu Xia}, \bibinfo{person}{Jinxuan Xu}, \bibinfo{person}{Yuqian Zhang}, {and} \bibinfo{person}{Hang Liu}.} \bibinfo{year}{2025}\natexlab{}.
\newblock \showarticletitle{A Survey of Uncertainty Estimation Methods on Large Language Models}. In \bibinfo{booktitle}{\emph{Findings of the Association for Computational Linguistics, {ACL} 2025, Vienna, Austria, July 27 - August 1, 2025}} \emph{(\bibinfo{series}{Findings of {ACL}})}. \bibinfo{publisher}{Association for Computational Linguistics}, \bibinfo{pages}{21381--21396}.
\newblock


\bibitem[Yang et~al\mbox{.}(2025b)]%
        {yang2024unifying}
\bibfield{author}{\bibinfo{person}{Liu Yang}, \bibinfo{person}{Fabian Paischer}, \bibinfo{person}{Kaveh Hassani}, \bibinfo{person}{Jiacheng Li}, \bibinfo{person}{Shuai Shao}, \bibinfo{person}{Zhang~Gabriel Li}, \bibinfo{person}{Yun He}, \bibinfo{person}{Xue Feng}, \bibinfo{person}{Nima Noorshams}, \bibinfo{person}{Sem Park}, \bibinfo{person}{Bo Long}, \bibinfo{person}{Robert~D. Nowak}, \bibinfo{person}{Xiaoli Gao}, {and} \bibinfo{person}{Hamid Eghbalzadeh}.} \bibinfo{year}{2025}\natexlab{b}.
\newblock \showarticletitle{Unifying Generative and Dense Retrieval for Sequential Recommendation}.
\newblock \bibinfo{journal}{\emph{Trans. Mach. Learn. Res.}}  \bibinfo{volume}{2025} (\bibinfo{year}{2025}).
\newblock


\bibitem[Yang et~al\mbox{.}(2025a)]%
        {yang2025sparse}
\bibfield{author}{\bibinfo{person}{Yuhao Yang}, \bibinfo{person}{Zhi Ji}, \bibinfo{person}{Zhaopeng Li}, \bibinfo{person}{Yi Li}, \bibinfo{person}{Zhonglin Mo}, \bibinfo{person}{Yue Ding}, \bibinfo{person}{Kai Chen}, \bibinfo{person}{Zijian Zhang}, \bibinfo{person}{Jie Li}, \bibinfo{person}{Shuanglong Li}, {and} \bibinfo{person}{Liu Lin}.} \bibinfo{year}{2025}\natexlab{a}.
\newblock \showarticletitle{Sparse Meets Dense: Unified Generative Recommendations with Cascaded Sparse-Dense Representations}. In \bibinfo{booktitle}{\emph{Advances in Neural Information Processing Systems}}, \bibfield{editor}{\bibinfo{person}{Danielle Belgrave}, \bibinfo{person}{Cheng Zhang}, \bibinfo{person}{Huan Lin}, \bibinfo{person}{Razvan Pascanu}, \bibinfo{person}{Piotr Koniusz}, \bibinfo{person}{Marzyeh Ghassemi}, {and} \bibinfo{person}{Niao Chen}} (Eds.), Vol.~\bibinfo{volume}{38}. \bibinfo{publisher}{Curran Associates, Inc.}, \bibinfo{pages}{93746--93770}.
\newblock


\bibitem[Zeghidour et~al\mbox{.}(2022)]%
        {zeghidour2021soundstream}
\bibfield{author}{\bibinfo{person}{Neil Zeghidour}, \bibinfo{person}{Alejandro Luebs}, \bibinfo{person}{Ahmed Omran}, \bibinfo{person}{Jan Skoglund}, {and} \bibinfo{person}{Marco Tagliasacchi}.} \bibinfo{year}{2022}\natexlab{}.
\newblock \showarticletitle{Soundstream: An end-to-end neural audio codec}.
\newblock \bibinfo{journal}{\emph{IEEE/ACM Transactions on Audio, Speech, and Language Processing}}  \bibinfo{volume}{30} (\bibinfo{year}{2022}), \bibinfo{pages}{495--507}.
\newblock


\bibitem[Zhang et~al\mbox{.}(2026)]%
        {zhang2025tokur}
\bibfield{author}{\bibinfo{person}{Tunyu Zhang}, \bibinfo{person}{Haizhou Shi}, \bibinfo{person}{Yibin Wang}, \bibinfo{person}{Hengyi Wang}, \bibinfo{person}{Xiaoxiao He}, \bibinfo{person}{Zhuowei Li}, \bibinfo{person}{Haoxian Chen}, \bibinfo{person}{Ligong Han}, \bibinfo{person}{Kai Xu}, \bibinfo{person}{Huan Zhang}, \bibinfo{person}{Dimitris~N. Metaxas}, {and} \bibinfo{person}{Hao Wang}.} \bibinfo{year}{2026}\natexlab{}.
\newblock \showarticletitle{Tok{UR}: Token-Level Uncertainty Estimation for Large Language Model Reasoning}. In \bibinfo{booktitle}{\emph{The Fourteenth International Conference on Learning Representations}}.
\newblock


\end{thebibliography}

\end{document}